\documentclass[letterpaper, 10 pt, conference]{ieeeconf}  

\IEEEoverridecommandlockouts                              

\usepackage{graphicx} 
\usepackage{mathptmx} 
\usepackage{amsmath} 
\usepackage{amssymb}  

\usepackage{multirow}
\usepackage{tabularx}

\usepackage{caption}
\usepackage{subcaption}
\usepackage{hyperref}
\usepackage{float}

\RequirePackage[normalem]{ulem}
\RequirePackage{color}\definecolor{RED}{rgb}{1,0,0}\definecolor{BLUE}{rgb}{0,0,1}

\title{\LARGE \bf
MAVNet: an Effective Semantic Segmentation Micro-Network for MAV-based Tasks}

\author{Ty Nguyen$^{1}$, Shreyas S. Shivakumar$^{1}$, Ian D. Miller$^{1}$, James Keller$^{1}$, Elijah S. Lee$^{1}$, Alex Zhou$^{1}$, Tolga \"Ozaslan$^{1}$\\
  Giuseppe Loianno$^{2}$, Joseph H. Harwood$^{3}$, Jennifer Wozencraft$^{3}$, Camillo J. Taylor$^{1}$, Vijay Kumar$^{1}$
\thanks{This work was supported by the MAST Collaborative Technology Alliance - Contract No. W911NF-08-2-0004, ARL grant W911NF-08-2-0004, ONR grants N00014-07-1-0829, N00014-14-1-0510, ARO grant W911NF-13-1-0350, NSF grants IIS-1426840, IIS-1138847, DARPA grants HR001151626, HR0011516850, and supported in part by the Semiconductor Research Corporation (SRC)
and DARPA.}
\thanks{
T. \"Ozaslan acknowledges the fellowship from The Republic of Turkey
Ministry of National Education.}
\thanks{$^{1}$ The authors are with the
GRASP Lab, University of Pennsylvania, Philadelphia, PA 19104 USA. {
        {\tt\footnotesize email: \{tynguyen, sshreyas , iandm,  jfkeller, elslee, alexzhou, ozaslan, cjtaylor, kumar\}}@seas.upenn.edu}}%
\thanks{$^{2}$ The author is with the New York University, Tandon School of Engineering, 6 MetroTech Center, 11201 Brooklyn NY, USA. 
{\tt\footnotesize email: \{loiannog\}@nyu.edu.}}
\thanks{$^{3}$ The authors are with the United States Army Corps of Engineers, Washington, DC 20314 USA. {
        {\tt\footnotesize email: \{joseph.h.harwood, jennifer.m.wozencraft\}}@usace.army.mil}}%
}

\begin{document}

\maketitle
\thispagestyle{empty}
\pagestyle{empty}

\begin{abstract}
Real-time semantic image segmentation on platforms subject to size, weight and power (SWaP) constraints is a key area of interest for air surveillance and inspection.  In this work, we propose MAVNet: a small, light-weight, deep neural network for real-time semantic segmentation on micro Aerial Vehicles (MAVs). MAVNet, inspired by ERFNet \cite{romera.toits2018}, features 400 times fewer parameters and achieves comparable performance with some reference models in empirical experiments.  Our model achieves a trade-off between speed and accuracy, achieving up to 48 FPS on an NVIDIA 1080Ti and 9 FPS on the NVIDIA Jetson Xavier when processing high resolution imagery.  Additionally, we provide two novel datasets that represent challenges in semantic segmentation for real-time MAV tracking and infrastructure inspection tasks and verify MAVNet on these datasets.
Our algorithm and datasets are made publicly available.
\end{abstract}


\section{Introduction}
\label{sec:introduction}
Autonomous MAVs capable of real-time, on-board image semantic segmentation can provide an effective solution for the target tracking problem in surveillance systems and the active sensing problem in inspection systems. Thanks to their high agility, MAVs are suitable for detecting and tracking moving targets, including targets that are relatively small and/or difficult to detect using standard technologies such as radar.  
Detecting and tracking these targets in real-time, on board is useful in 
1) real-time semantic mapping for inspection and surveillance; 2) real-time detection and classification of other MAVs for formation control; 3) real-time detection of other MAVs for privacy and security. 
However, this problem is challenging since MAVs can appear in a wide variety of orientations and distances against a variety of backgrounds. Additionally, the SWaP constraints of the deployment platform itself impose severe constraints on computational capability. 

In addition, MAVs equipped with on-board sensors and computers can be a viable and inexpensive solution for assisting humans with complex, labor-intensive, high-risk tasks. Some typical examples are the periodic inspection and maintenance of critical infrastructure such as dams and penstocks~\cite{ozaslan_kumar.ral18}, or monitoring wildfire. Real-time semantic segmentation in these cases, where communication bandwidth is often limited, enables active sensing modalities, where the robot autonomously explores and investigates depending on what it currently senses from the environment.  

In the field of semantic segmentation, deep learning has become the \textit{de-facto} approach with superior accuracy and robustness over classical machine learning approaches \cite{lecun.nature15, deng.now14}.
Some prominent application areas are agricultural inspection for fruit counting \cite{sa_mccool.sensors16}, disease detection \cite{mohanty_salathe.fps16}, vehicle and pedestrian traffic monitoring \cite{sermanet_lecun.cvpr13, lv_wang.its15}, and structural health monitoring of critical infrastructure \cite{cha_buyukozturk.cacie17, yeum2015vision, makantasis2015deep}. However, most evaluation of semantic segmentation algorithms is performed on datasets like KITTI \cite{geiger.cvpr2012} or CityScapes \cite{cordts.cvpr2016}, which are useful for self-driving cars, but are unrepresentative of the challenging environments encountered by MAVs or the complex environments in the aforementioned application areas. To fill this gap, we first provide two novel datasets 1) the MAV segmentation dataset used for evaluating MAV segmentation; 2) the penstock dataset used for evaluating corrosion segmentation in penstocks. Images in these datasets, captured by on-board cameras which can be either RGB or grayscale, reflect real-world sensing constraints and are often noisy with poor illumination.



\begin{figure}[t]
	\centering
	\includegraphics[width=\linewidth]{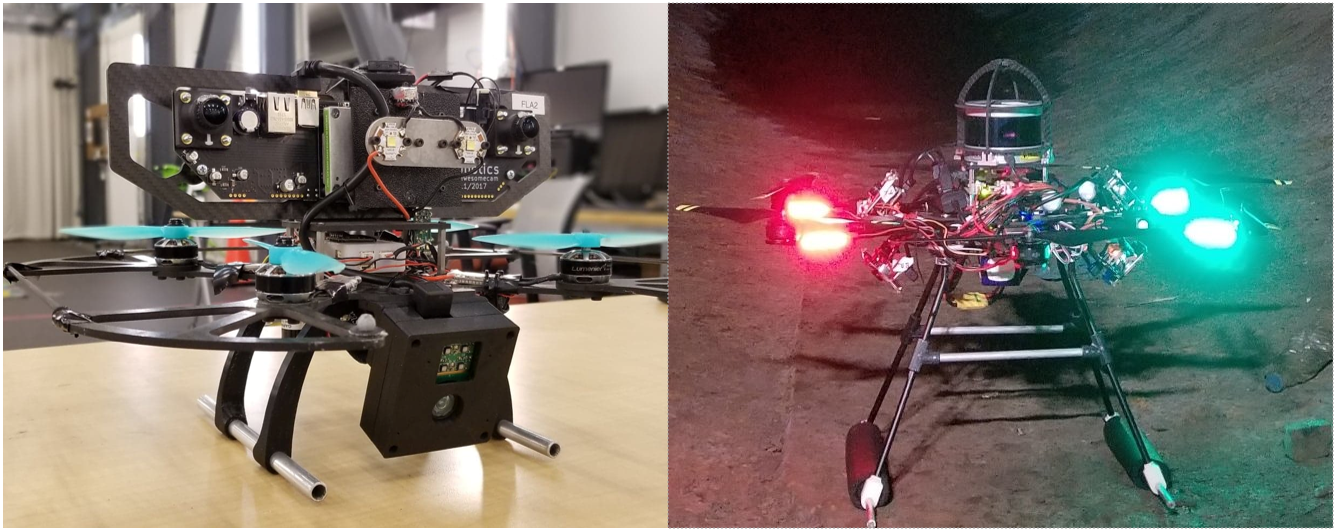}
	\caption{
	    Left: the Falcon 250, featuring a NVIDIA Jetson TX2, used in MAV segmentation; Right: the MAV equipped with four on-board cameras $\&$ LEDs, has been developed to inspect penstocks in dams for hydroelectric power. }
	\label{fig:drone}
	\vspace{-3mm}
\end{figure}


Despite the successes in classification and segmentation, deep learning approaches often require significant computational power. Since the introduction of residual networks such as ResNet \cite{he.cvpr2016} which add increasing numbers of layers to the network, there has been much emphasis on accuracy over compactness and efficiency of algorithms. 
While moving complex processing to the cloud is a common solution for constrained edge devices, it inhibits the MAV's operations in remote areas or subterranean regions where wireless connectivity is often not available.  

Examples of such applications include relative visual localization of other robots in multi-robot systems \cite{saska.icuas2015}, semantic mapping \cite{maturana.irs2017}, and damage detection for infrastructure inspection \cite{rau2017bridge, ozaslan_kumar.ral17}.  There are some recent research in deep learning inference at the edge \cite{romera.toits2018, hochstetler.sec2018}.  However, most of these works focus on datasets collected from ground vehicles, which do not suffer from SWaP constraints, and lack the challenges presented with MAV imagery.

In short, our primary contributions are as follows. 
\textit{First}, we publicly release two challenging datasets publicly to encourage further work in developing algorithms for real-time on-board semantic segmentation in challenging environments. 
\textit{Second}, we present a novel deep learning network for achieving high speed image segmentation at full resolution with minimal resources and computational demands. 
\textit{Third}, we evaluate the network on two datasets, demonstrating the flexibility of our approach as well as performance improvements over the current state-of-the-art. 
\section{Related Work}
\label{sec:related_work}
\subsection{Real-Time Deep Learning}
State-of-the-art segmentation algorithms such as ResNet \cite{he.cvpr2016} and VGG \cite{simonyan.2014} have achieved excellent performance on a variety of different datasets.
However, these are highly complex networks, involving many layers, and require powerful processors for inference at high speed.
In \cite{enet_paszke2016}, the authors present ENet, which can run at up to 7 fps on a Jetson TX1, but only with low resolution input images.
ErfNet \cite{romera.toits2018} builds on ENet, achieving superior accuracy by using a more complex but slightly slower architecture.  
The authors of MobileNet \cite{mobilenets_howard2017} develop an architecture with several hyperparameters allowing the user to tune their model for particular constraints required by the application.  
ESPNet \cite{espnet_mehta2018} uses efficient spatial pyramids to achieve accuracy close to ResNet but at higher inference speeds.  
The authors of ICNet \cite{ICNet_Zhao_2018} report that they significantly outperform ENet on CityScapes while running at 30fps with $1024 \times 2048$ resolution, but this performance is limited to desktop grade GPUs such as the Titan X. Furthermore, they do not test their algorithm on embedded devices such as the Jetson.

While these methods have achieved accurate results at reasonably high inference rates on a large number of classes, validation for all of these methods is typically performed on driving datasets such as KITTI or Cityscapes.


\subsection{Deep Learning for Visual Inspection}


There has been significant interest in using deep learning techniques for infrastructure inspection.
In a recent study, \cite{cha_buyukozturk.cacie17} introduces a sliding-window technique using a CNN-based classifier to detect cracks on concrete and steel surfaces. 
The major drawback of this method is that it cannot satisfy real-time processing requirements and would fail in detecting small defects which are very frequent in our images. 
This type of framework is also not data-efficient since it processes an image patch as a single sample. 

In \cite{park2016machine}, the authors use CNNs to detect defects in different types of materials such as fabric, stone and wood.
They compare their network to classical machine learning methods such as the Gabor Filter, Random Forest and Independent Component Analysis.
However, the authors do not optimize the inference speed of their classifier.

In a similar application to ours, \cite{makantasis2015deep} propose to feed a CNN with low-level features such as edges so as to obtain a mixture of low and high level features before classifying pixels to detect defects on concrete tunnel surfaces. 
Unlike this work, we propose an end-to-end fully convolutional neural network that does not depend on handcrafted features and also works with arbitrary input image sizes. 
In fact, some of the low-level features used in~\cite{makantasis2015deep} are neither easy to obtain nor provide useful information for the CNN such as edges, texture and frequency.
This is paricularly true for the noisy, dusty, and poorly lit images captured in our challenging datasets, where standard edge detection will often not provide useful information, only finding noise or dust trails.

\section{Datasets}
\label{sec:dataset}

In this study, we evaluate the performance of deep network models on two different datasets that we collected using autonomous MAVs.

\begin{figure*}[t]
  \vspace{3mm}
    
     \begin{minipage}{.45\textwidth}
     \raggedright
     \centering
     \subcaption{}
      \vspace{-2mm}
       \includegraphics[width=\linewidth]{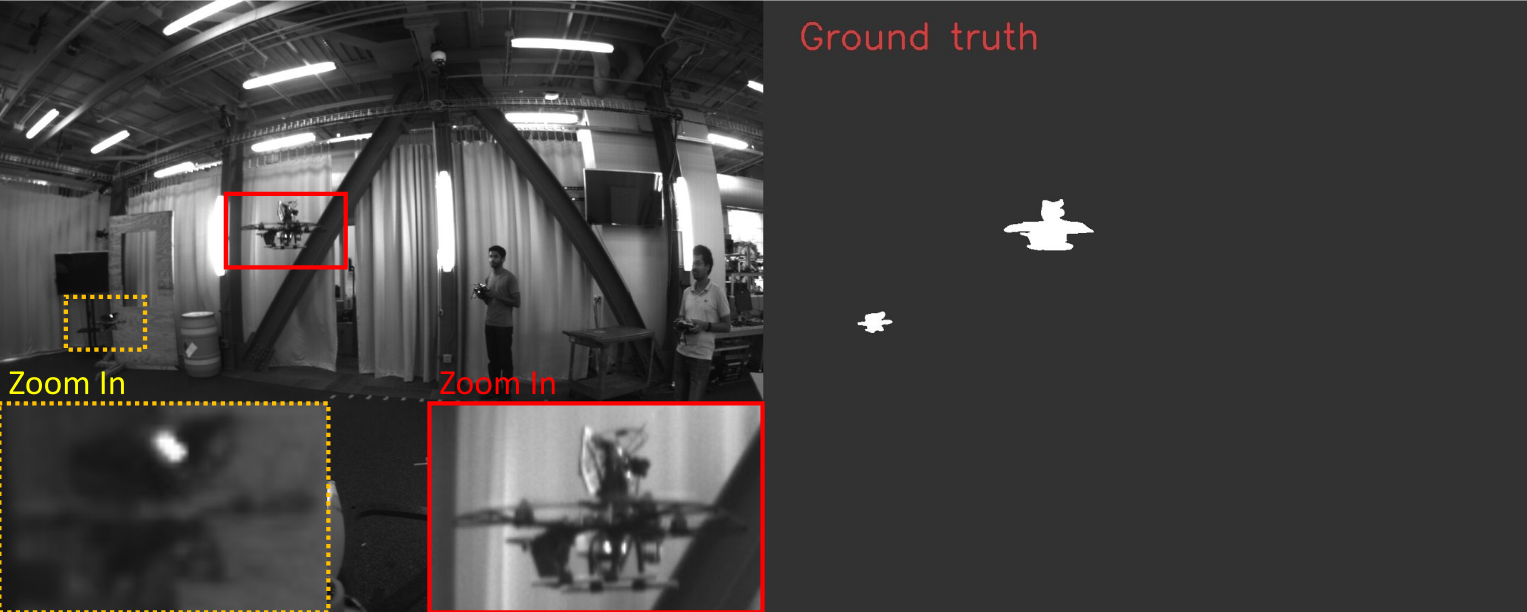}
    \end{minipage}  
    \hfill
    \begin{minipage}{.45\textwidth}
      \raggedleft  
      \subcaption{}
      \vspace{-2mm}
      \includegraphics[width=\linewidth]{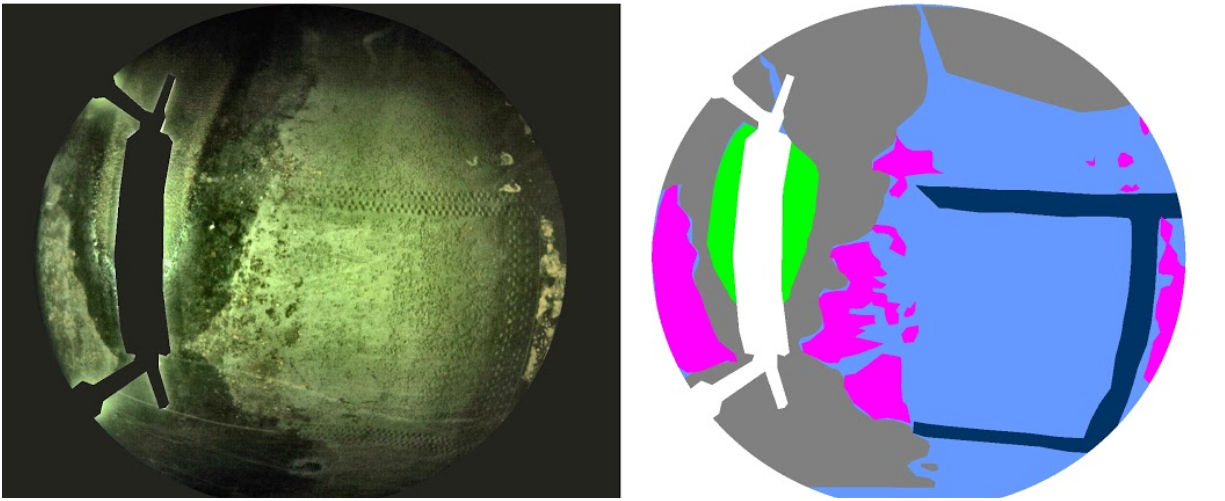}
    \end{minipage} \\

    \caption{
        (a) A sample image from the drone dataset. 
        From left to right: input image,  labeled image. White: drone, gray: background
        (b) A sample image captured by one of the fish-eye cameras from the penstock dataset. 
        From left to right: input image, labeled image. 
        Pink: corrosion, light blue: background, dark blue: rivet, green: water, gray: ignore; 
    }
    \label{fig:intro_image}
    \vspace{-3mm}
\end{figure*}

\subsection{MAV Segmentation Dataset}

The first dataset (Fig.~\ref{fig:intro_image}a) is a multi-robot flying dataset collected indoors, whose primary purpose is to train real-time vision-guided MAV tracking systems with the goal of controlling swarms of MAVs without explicitly communicating state information to neighboring vehicles. Fig.~\ref{fig:intro_image}a shows an image sample and its corresponding labels. The images are captured by an Open Vision Computer~\cite{quigley2018open} which are integrated in our custom Falcon 250 MAV platforms, and feature an NVIDIA Jetson TX2 alongside gray-scale Python-1300 cameras. This is a challenging dataset as the relative motion between the MAVs is constantly changing, resulting in a large variance in the size of the target objects with respect to the background of the image. 

The training dataset consists of the original images, along with around $1000$ pixel-wise labels sampled from $12$ flights of $3$ MAVs within an indoor area. 

For testing, we collect and label another $\sim 300$ images sampled from a video captured in a different, more cluttered, indoor location. We also slightly modify the target MAV to detect from the ones appeared in the training dataset by removing the sensors from the robot, thereby testing the resilience of models to modifications to the target robot. 


\subsection{Penstock Dataset}
The second dataset provided in this study is collected using a customized DJI-F550 MAV described in \cite{ozaslan2017autonomous} that autonomously flies inside a penstock at Center Hill Dam, TN. 

There are four fish-eye cameras mounted on the MAV such that the combined field of view covers the annulus of the tunnel. 
We provide two sequences of images taken from two flights with different lighting conditions, using one for training and the other for testing. 


We apply limited adaptive histogram equalization using CLAHE from OpenCV 4.0 to mitigate the brightness imbalance with a clip limit of $2$ and the tile grid size of $8$. Also, image regions occluded by the MAV's landing gear, camera lens covers, and propellers are masked out.



We use $4$ classes to label pixels: background, corrosion, rivet, and water, as shown in Fig.~\ref{fig:intro_image}b. Expert labellers are required to precisely label this challenging dataset. Furthermore, each image is separately labelled by \textit{three} experts. If there is a disagreement about a label instance, the labellers discuss and vote. If there is no conclusive agreement at this point, the instance is labelled as ignore.   
We divide images between training and test sets as follows. Images captured from the two cameras on the left side of the MAV from the first sequence are sampled and labeled for the training set. Images captured from the two cameras on the right side of the MAV from the second sequence are for testing. The rest are used for the validation set. This way, no part of the penstock seen in training is seen by the classifier when testing. 

Unlike the MAV dataset, this dataset is relatively small since the image sequences are quite short, due to the limited length of the penstock where the MAV flies to collect data. Additionally, labelling these images is 1) significantly more complex and time consuming than in the MAV dataset; and 2) requiring more area experts. 
In fact, the training set consists of $39$ images, the validation set consists of $64$ images and the test set consists of $35$ images. However, this dataset poses different challenges from the MAV dataset. Moreover, evaluating the deep network models on both datasets provide a complete view about the effectiveness of the models on large and small training datasets and very different classification contexts.  
\section{Network Design}
\label{sec:network_design}

Unlike common state-of-the art deep network models benchmarked on large datasets such as Cityscapes and MS COCO, networks intended for robotic vision tasks run on on-board processors and must therefore satisfy performance, computational, and memory constraints. 
As we will demonstrate in our experiments and others have observed \cite{romera.toits2018}, it is insufficient to merely reduce the number of parameters in a more complex network, particularly without making significant performance sacrifices.  
We therefore believe that designing a new network structure, rather than attempting to re-scale existing ones, is necessary. Inspired by ErfNet~\cite{romera.toits2018}, this section details the intuitions and experiments that lead to our proposed network design.  

\subsection{Downsampling}

Despite downsampling having undesired side-effects such as spatial information loss and the introduction of potential checkerboard artifacts during upsampling to the original image size, it is still a widely utilized step for a variety of reasons.
Downsampling can help reduce spatial redundancy, making the precedent filtering layers operate with lower resolution features and thereby saving significant computational and memory cost. 
Additionally, features obtained from filtering using the reduced input have a larger receptive field, making them able to gather information from over a broader context. 
This capability is essential for detecting objects with significant variation in size.  


A simple trick to mitigate the first side effect of downsampling ---  spatial information loss --- is to increase the number of output features by a factor equal to the downsampling factor. FCN~\cite{fcn_long2015fully} and UNet~\cite{unet_ronneberger2015u} go a step further and utilize skip connections from early layers of the encoder to the decoder to preserve spatial information. However, these long skip connections require a large amount of memory to transfer the intermediate results from the encoder to the decoder. SegNet~\cite{segnet_badrinarayanan2017} and ENet~\cite{enet_paszke2016} solve this problem by memorizing the elements chosen in the maxpooling step (in the encoder) to utilize in the downsampling process (in the decoder).

We investigate the effect of downsampling by evaluating two variants of ERFNet: one with downsampling and one without downsampling on our two datasets. Results show that the later version without downsampling does not significantly perform better than the former version while requiring more time and memory in inference. We also find that the downsampler block used in~\cite{romera.toits2018} results in inferior performance compared to conv-conv-pool in~\cite{simonyan2014very} when the number of layers and number of features of the network are significantly shrunk. Thus, we make use of two conv-conv-pool blocks to downsample the input image as the first two blocks of our network.

\subsection{Dilated Convolution}
Dilated convolution, or atrous convolution, is an effective and inexpensive way to increase the receptive field of a network as demonstrated in successive works such as~\cite{romera.toits2018, chen2018encoder}. By stacking multiple dilated convolution layers with a dilation rate of more than one, the preceding features can achieve an exponential increase in the receptive field given the same number of parameters and computations that regular convolutions use.  Intuition suggests that a larger receptive field allows the network to see not only the object but also the context in which the object stands. This context can be leveraged to improve segmentation performance. However, unlike ErfNet, we find that stacking more than $K$ dilated convolutions with a dilation rate of $2$ (where $2^K$ is equal to input image size) is unnecessary as the preceding dilated convolutions have no effect. This observation informs our removal of a significant number of dilated convolution layers from ErfNet.

\subsection{Depth-wise Feature Aggregation Block (DWFab)}
\begin{figure}
    \centering
    \includegraphics[width=0.9\linewidth]{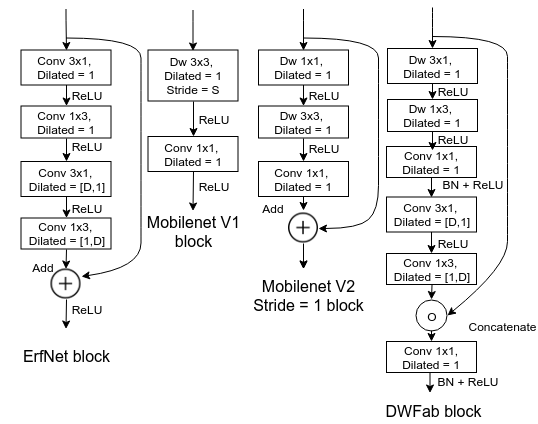}
    \caption{Diagrams of different Conv blocks. The ErfNet Conv block and DWFab block utilize dilated convolution with factor D $> 1$.}
    \label{fig:non_bottle_v_depth}
\end{figure}

The backbone of our MAVNet is a simple but effective depth-wise feature aggregation block (DWFab). Fig.~\ref{fig:non_bottle_v_depth} sketches the DWFab's structure in comparison with the convolution blocks (conv block) used in Mobilenet v1~\cite{mobilenets_howard2017}, Mobilenet v2~\cite{sandler2018mobilenetv2} and ErfNet~\cite{romera.toits2018}. Compared to conv blocks used in Mobilenets, DWFab and ErfNet blocks both utilize dilated convolutions to efficiently increase the receptive field and handle object size variance in the input images. We investigate the effect of the dilated convolutions by conducting an experiment with MAVNet in which all dilation factors are set to $1$. This modified MAVNet does not converge when training on the MAV dataset. On the penstock dataset, it achieves an IoU of $31\%$ compared to $36.24\%$ on the original MAVNet. We conclude that dilated convolution helps improve performance. 

The main difference between our DWFab block and the conv block used in ErfNet is that the first two conv in the DWFab block use depth-wise separable convolutions followed by a $1 \times 1$ convolution. Theoretically, this alternative can achieve about $2$ times speedup but in practice, it can be slower when using the $separable\_conv2D$ function in Tensorflow.     


\subsection{Network Architecture} 
\begin{figure}
    \centering
    \includegraphics[width=0.95\linewidth]{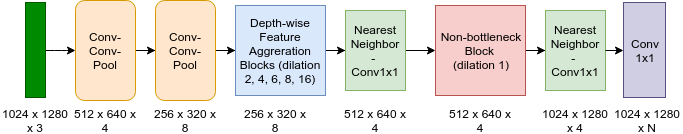}
    \caption{Network Architecture. Input used in this study is of size $1024 \times 1280 \times 3$ and the output segmentation is $1024 \times 1280 \times \textbf{N}$, with \textbf{N} is the number of classes.}
    \label{fig:network_design}
    \vspace{-2mm}
\end{figure}

As can be seen in Fig.~\ref{fig:network_design}, our network is quite simple compared to ErfNet, UNet and ENet. The encoder part consists of two Conv-Conv-Pool blocks, followed by four DWFab blocks that have dilation rates of $2, 4, 8, 16$ respectively.   

The decoder consists of two upsampling blocks with a non-bottleneck, as used in ErfNet, in between, and a $1\times 1$ convolution at the end to output the logits. These umpsampling blocks differ from the upsamplers used in ErfNet in which deconvolution is replaced by nearest-neighbor upsampling followed by regular convolution. This replacement helps mitigate the checkerboard issue caused by deconvolution~\cite{odena2016deconvolution}.

Such short decoder is used since it can significantly reduce computation as well as eliminate the need for long skip connections, thanks to the shallow network. We empirically found that a long decoder without a long skip connection is harder to train, resulting in inferior performance. 


To demonstrate the effectiveness of the DWFab block, in the following sections, we carry out experiments to compare MAVNet with UNet, ENet, and ErfNet. In addition, we create a variant of ErfNet with the same number of blocks as that of MAVNet. This network is referred to as S-ErfNet for the remainder of this paper.

\section{Training}
\label{sec:training}
\subsection{Focal Loss for Multi-class Classification}
Class imbalance is a common problem which often appears when performing multi-class segmentation.
In robotics and medical imaging applications, class imbalance is exacerbated by small training sets, due to the difficulty and cost involved in gathering data as well as the need for expert labellers. 

To mitigate this problem, we make use of the focal loss, introduced in \cite{lin2018focal}. We can generalize focal loss for the multi-class classification problem as follows:  
\begin{equation}
\mathbf{L}_{Focal} = - \frac{1}{\mathbf{N}} \sum_{n=1}^\mathbf{N} \sum_{c=1}^\mathbf{C} (1-\hat{y}_{nc})^\gamma y_{nc} \log{\hat{y}_{nc}},
\label{eq:L_focal}
\end{equation}
where $\gamma > 0$ is a tunable parameter. The effect of the focal loss and $\gamma$ value can be understood in the following manner:
When a difficult sample is misclassified, with the true class given low confidence ($\hat{y}_{nc}$ is small), the weighting factor becomes close to $1$, preserving that sample's contributions to the total loss. 
In contrast, an easy sample correctly classified with a high confidence value ($\hat{y}_{nc}$ is large), will have its weight close to $0$, reducing its contribution to the total loss. 
In summary, the focal loss function can appreciate the weighting of difficult samples, regardless of which class they belong to, by giving more them more weight and depreciating easy samples. 
In our experiments, we set $\gamma = 2$ as recommended by the authors of \cite{lin2018focal}. Since the MAV dataset presents a huge imbalance between positive samples  (MAV pixels) and negative samples (background pixels), we introduce additional weights for each sample. Eq.~\ref{eq:L_focal} becomes
\begin{equation}
\mathbf{L}_{Focal} = - \frac{1}{\mathbf{N}} \sum_{n=1}^\mathbf{N} \sum_{c=1}^\mathbf{C}w_c (1-\hat{y}_{nc})^\gamma y_{nc} \log{\hat{y}_{nc}},
\label{eq:L_w_focal}
\end{equation}
where $w_c$ is the corresponding weight for class $c$. Empirically, we set $w_{\text{MAV}} = 20$, and $w_{\text{background}} = 1$.

We investigate the effectiveness of focal loss by comparing the performance of MAVNet trained using focal loss with that of MAVNet trained using cross-entropy loss. Results are shown in Sec.~\ref{sec:benchmarks}. For simplicity, from now on, we refer to the MAVNet model trained using focal loss whenever a loss function is not explicitly mentioned. 

\subsection{Training Scheme}
All the deep network models investigated in this study are implemented in Tensorflow \cite{abadi2016tensorflow}. The training procedure is the same with all models: use mini-batch gradient descent with a batch-size of 4 and the Adam optimizer~\cite{kingma2014adam} with $\beta_1 = 0.9$,
$\beta_2 = 0.999$, $\epsilon = 10^{-8}$, and learning rate $= 0.001$. All models are trained until convergence, and the loss fuction ceases to decrease. Online data augmentation is used including random rotation, random cropping and padding, random gamma shifting, random brightness shifting, and random color shifting. Our implementation is publicly available at  \href{link}{https://github.com/tynguyen/MAVNet}.  
 
\section{Benchmarks}
\label{sec:benchmarks}
We benchmark our proposed network in comparison with networks including UNet, ErfNet, ENet and S-ErfNet on both the MAV and penstock datasets. The performance is evaluated using different metrics as follows. 

\subsection{Metrics}
For each model, we report three metrics commonly used in semantic segmentation: Intersection over union (IoU), false negative (FN) rate, and false positive (FP) rate. 

\begin{align}
  \text{IoU} &= \frac{TP}{TP + FP + FN}, \\ 
  \text{False Negative Rate} &= \frac{FN}{FN + TP}, \\
  \text{False Positive Rate}   &= \frac{FP}{FP + TN},
\end{align}
where $TP=$ Pixels correctly classified as the object by the classifier; $FP=$ Pixels not classified as the object in the ground truth, but classified as the object by algorithm; $TN=$ Pixels not classified as the object in ground truth and by algorithm; $FN=$ Pixels classified as object in ground truth, but not classified as object by algorithm. It is desirable to obtain a segmentation model with a high IoU and low FP and FN rates. However, it is often often infeasible to design such model in practice.  Instead, there is often a trade-off between these metrics when selecting a model. 

In addition, we introduce the centroid distance metric for the MAV dataset which is calculated as the $\mathbf{L2}$ norm of the difference between the centroid of the MAV detected and that of the MAV in the ground truth, in pixels.  For the tracking application, where the objective is simply to localize the target, this is the most relevant performance metric.

\subsection{MAV Segmentation Dataset}
Table~\ref{tab:drone_performance} details the four metrics for the MAV object class on the MAV segmentation dataset. As the table demonstrates, ENet has the best IoU values, ErfNet has the best FN rate, UNet has the best FP rate, and MavNet the lowest centroid distance error.
MAVNet, while having an IoU lagging behind ENet, UNet, and ErfNet, has the second best value in FN, and the best centroid distance metric. Indeed, it has FN rate of $3.34\%$ compared to $1.97\%$ of the best and centroid distance error of $3.36$ compared to $3.46$ of the second best. Nevertheless, the superior centroid distance error gives MAVNet higher preference in practice where a real-time filtering algorithm is often used to solve the tracking problem. It is also worth noticing that that focal loss yields higher false positive rate than cross-entropy loss does while having better false negative rate.


 


\begin{table}[t]
\centering
\begin{tabular}{l ||c c c c}
& \multicolumn{4}{c}{MAV}   \\ \cline{2-5} 
    & IoU & FN Rate & FP Rate & Centroid Distance \\ \hline \hline
UNet& $63.50$& $14.27$& $\mathbf{0.14}$& $41.40 \pm 602.7$\\
ErfNet& $55.30$& $\mathbf{1.97}$& $0.31$& $4.61 \pm  1.81$\\
ENet& $\mathbf{67.62}$& $5.8$& $0.16$& $3.46 \pm 2.71$\\
S-ErfNet& $42.42$& $8.20$& $0.48$& $4.55 \pm 3.54$\\
MAVNet-Focal& $44.30$& $3.34$& $0.48$& $\mathbf{3.36 \pm 1.85}$\\
MAVNet-CE& $46.00$& $10.00$& $0.39$& $3.71 \pm 4.11$\\
\end{tabular}
\caption{Performance on MAV dataset. MAVNet is trained using focal loss (-Focal) and cross-entropy loss (-CE) }
\label{tab:drone_performance}
\vspace{-3mm}
\end{table}

\subsection{Penstock Dataset}
Tab.~\ref{tab:penstock_performance} details the three metrics for each class on the penstock dataset. 

As can be seen, the performance of each model varies with respect to each different class of objects. For example, UNet performs well on corrosion, but poorly on rivets. MAVNet while having an IoU of $57.66\%$ compared to the $61.67\%$ of Unet for corrostion, has better IoU for other classes. In addition, MAVNet has the lowest FN rate on corrosion and rivet segmentation. Overall, MAVNet has the highest average IoU over all classes. 
\begin{table*}[t]
\centering
\begin{tabular}{l || c c c||c c c||c c c|| c}
             & \multicolumn{3}{c}{Corrosion} & \multicolumn{3}{c}{Rivet}  & \multicolumn{3}{c}{Water}  \\ \cline{2-11} 
         & IoU & FN Rate & FP Rate  & IoU & FN Rate & FP Rate  & IoU & FN Rate & FP Rate & Mean IoU \\ \hline \hline
         
UNet& $\mathbf{61.67}$& $26.04$& $\mathbf{5.68}$& $0.00$& $100.00$& $\mathbf{0.00}$& $1.83$& $98.12$& $0.03$ & $21.18$\\
ErfNet& $35.84$& $56.73$& $6.98$& $31.68$& $64.70$& $0.6$& $5.90$& $92.70$& $0.23$& $24.47$\\
ENet& $45.12$& $43.00$& $7.86$& $34.07$& $62.55$& $0.52$& $0.00$& $100.00$& $\mathbf{0.00}$ & $26.40$\\
S-ErfNet& $50.49$& $28.24$& $12.33$& $42.73$& $53.72$& $0.42$& $2.90$& $96.57$& $0.26$ & $32.04$\\
MAVNet-Focal& $57.66$& $22.30$& $10.06$& $\mathbf{48.03}$& $\mathbf{41.43}$& $1.14$& $3.05$& $93.46$& $0.95$ & $\mathbf{36.24}$\\
MAVNet-CE& $52.20$& $\mathbf{17.21}$& $17.18$& $42.41$& $44.61$& $1.45$& $\mathbf{6.10}$& $\mathbf{90.16}$& $0.68$ & $33.56$\\
\end{tabular}
\caption{Class-wise performance on penstock dataset. All metrics are in $(\%)$, best values are in bold.}   
\label{tab:penstock_performance}
\vspace{-3mm}
\end{table*}

\section{Performance Analysis}
\label{sec:performance_analysis}
\subsection{Speed and Performance Tradeoff}
\label{sec:speed_vs_performance}
In this section, we report the performance of models with respect to inference speed and their model's complexity. Details are given in Tab.~\ref{tab:performance_vs_resources}, where performance in IoU is the average IoU over all classes on each dataset except the background. 

To measure the speed, we run the pretrained models and report the inference speed of models on three platforms 1) the NVIDIA Jetson Xavier, 2) the Falcon 250 that equipped with a NVIDIA Jetson TX2, and 3) the NVIDIA Jetson Nano. Speed is measured for input images of size $1024 \times 1280 \times 3$ with a batch size of $1$.

\begin{figure}[h]
\centering
\includegraphics[width=0.8\linewidth]{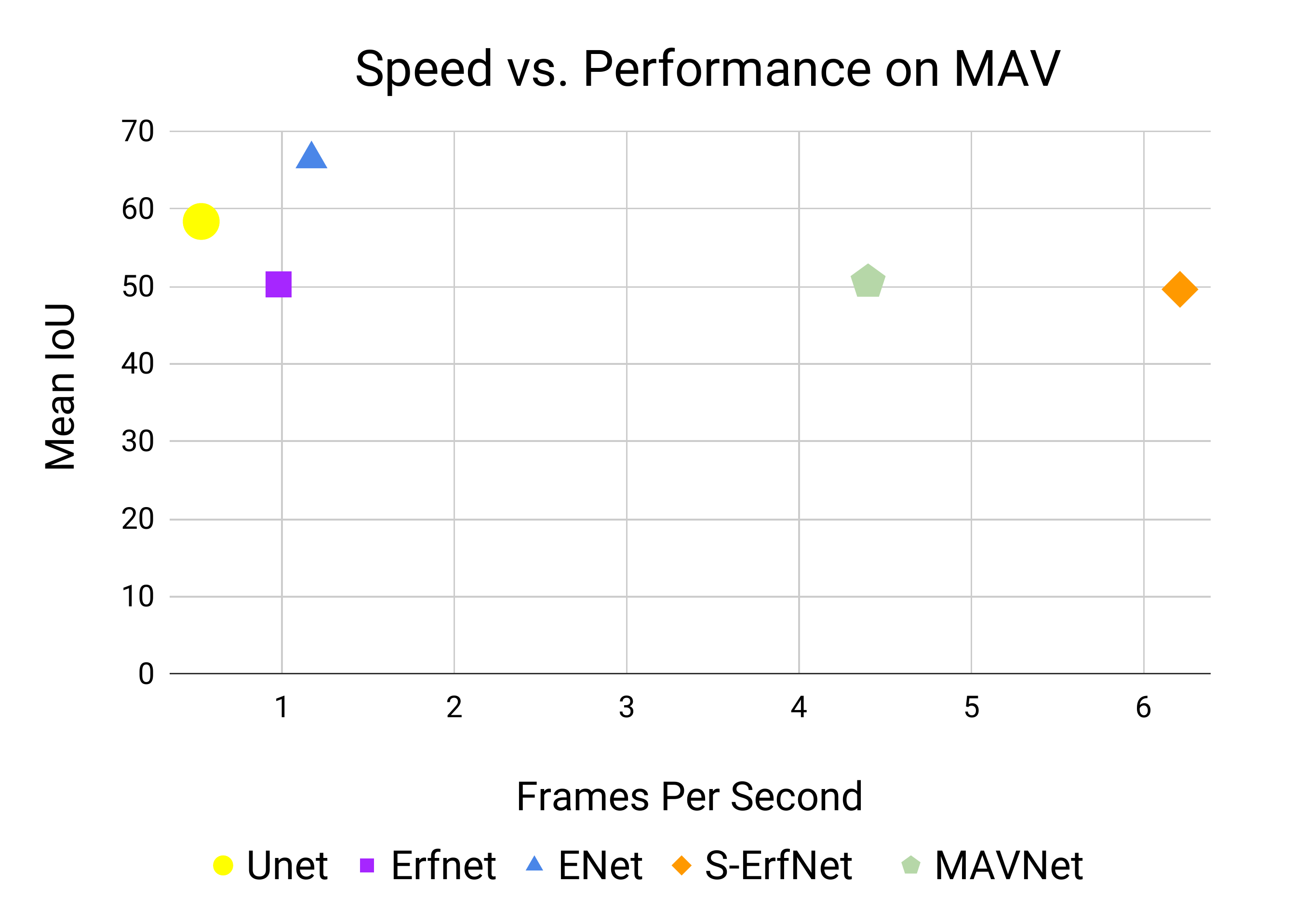}
\caption{Speed v.s. mean IoU over MAV object class on MAV dataset. MAVNet performance in IoU is inferior to UNet, ENet but runs much faster. Running time is measured on Falcon 250.}
\label{fig:drone_speed_vs_performance}
\end{figure}

\begin{figure}[h]
\centering
\includegraphics[width=0.8\linewidth]{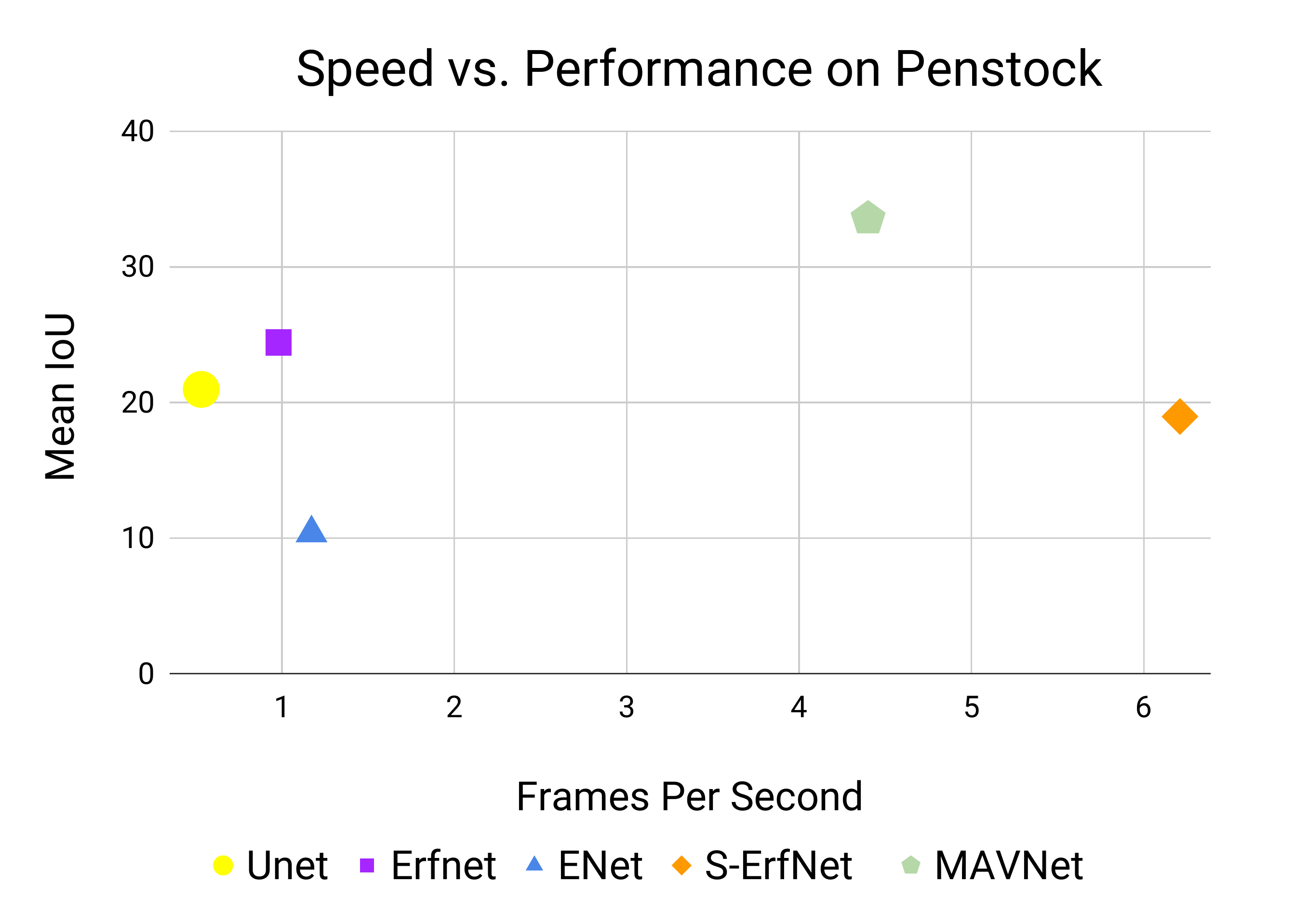}
\caption{Speed v.s. mean IoU over $3$ object classes on penstock dataset. MAVNet outperforms other methods while running $40\%$ slower than the fastest, S-ErfNet. Running time is measured on Falcon 250.}
\label{fig:penstock_speed_vs_performance}
\end{figure}

Fig.~\ref{fig:penstock_speed_vs_performance} and ~\ref{fig:drone_speed_vs_performance} visualize the corresponding inference speed and performance of each model in each dataset running on our Falcon 250. Note that all models are trained from scratch for each dataset using the same training scheme, and the trained models are not optimized using optimization techniques and tools such as TensorRT. These two figures show the effectiveness of our network design. 

As can be seen in both figures, MAVNet has a speed of $4.4$ fps compared to the fastest one, S-ErfNet which tops $6.2$ fps while having better performance. MAVNet even outperforms the original ErfNet and other methods which run $4-6$ times slower on the penstock dataset. Tab.~\ref{tab:performance_vs_resources} demonstrates MAVNet's flexibility, yielding consistent results for both datasets and its compactness. MAVNet, with around $4300$ parameters, has roughly $1800$ times fewer parameters than UNet and $400$ times fewer parameters than ErfNet. These advantages make it perfectly suitable for embedded systems and MAV tasks.  

\begin{table}[t]
\centering
\begin{tabular}{l ||c|c|c|c|c}
  Items          & UNet & ErfNet     & ENet    & S-ErfNet & MAVNet\\ \hline \hline
  IoU 1 ($\%$) & $63.50$ & $55.30$  & $67.62$  & $42.42$      & $44.30$\\  
  IoU 2 ($\%$) & $21.18$   & $24.47$  & $26.40$ & $32.04$      & $36.24$\\
  Nano (fps)    & $0.26$   & $0.41$    & $0.50$   & $3.43$    & $2.13$\\  
  Falcon (fps)    & $0.55$   & $0.97$    & $1.17$   & $6.21$    & $4.40$\\  
  Xavier (fps)    & $1.3$   & $2.1$    & $2.9$   & $13.6$    & $8.9$\\  
  Params (mils)  & $7.76$   & $2.06$    & $0.37$   & $0.0039$       & $0.0043$\\
  Avg Speedup        & $1\times$     & $1.7\times$    & $2.1\times$  & $11.6\times$  & $7.7\times$\\  
 
  Param savings  & $1\times$   & $3.7\times$    & $21\times$   & $1990\times$       & $1772\times$\\ \hline
  
\end{tabular}
\caption{Performance v.s. complexity comparison. Inference time is measured in fps, in Jetson Nano, Falcon 250 and Jetson Xavier platforms. The input image is $1024 \times 1280 \times 3$. IoU 1 is the average of IoU of classes on the MAV dataset.  IoU 2 is the average of IoU of classes on the penstock dataset.}
\label{tab:performance_vs_resources}
\vspace{-3mm}
\end{table}

\subsection{Discussion}
\label{sec:qualitative_results}
Tab.~\ref{tab:drone_performance} and~\ref{tab:penstock_performance} show that ErfNet, S-ErfNet and MAVNet tend to have higher FP rates while having lower FN rates compared to ENet and UNet. Since the high FP rate happens with MAVNet in two cases - using focal loss and cross-entropy loss - it is suggested that this phenomenon comes from the network structure of these models where a stack of dilated convolution is used to give high receptive field in the last layers, and there is a lack of long skip connections as in UNet. 

\begin{figure*}[t!]
    \vspace{3mm}
    \centering
    \includegraphics[width=0.8\textwidth]{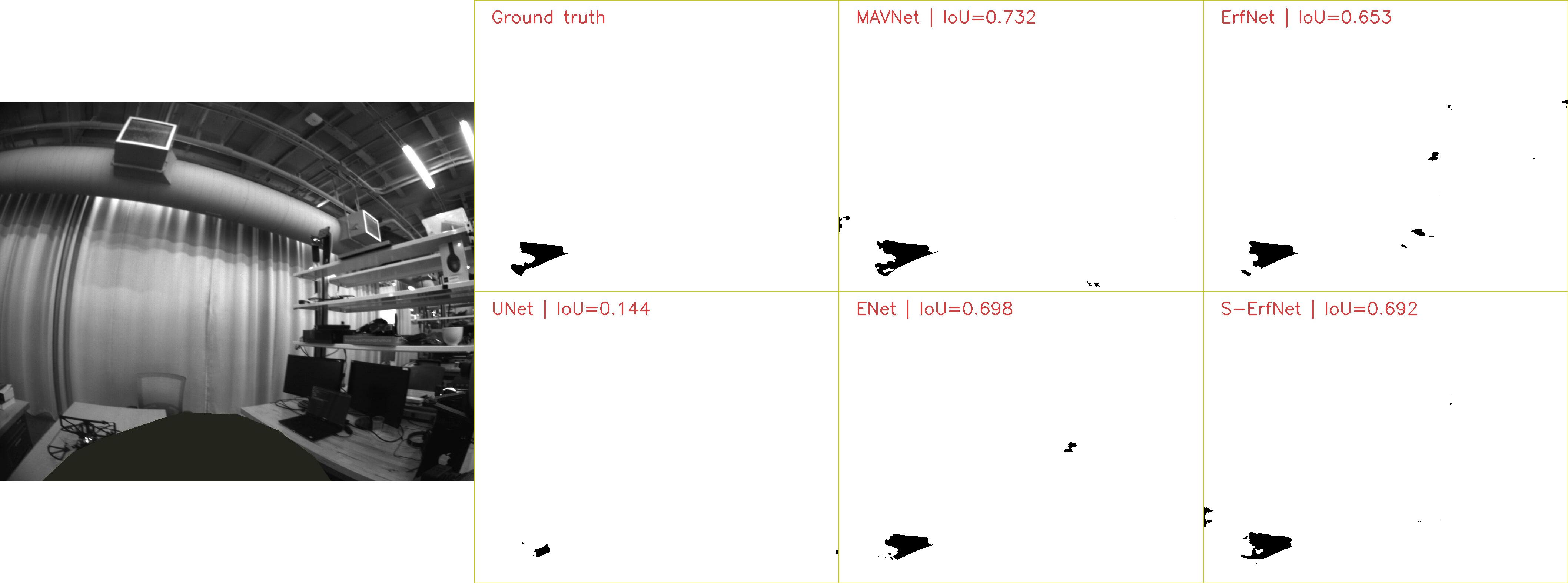}
    
    \caption{Success case in MAVNet dataset. UNet fails to detect the MAV. Colors: black - MAV, white - background}
    \label{fig:drone_qualitative_result}
\end{figure*}

\begin{figure*}[t!]
    \vspace{3mm}
    \centering
    \includegraphics[width=0.8\textwidth]{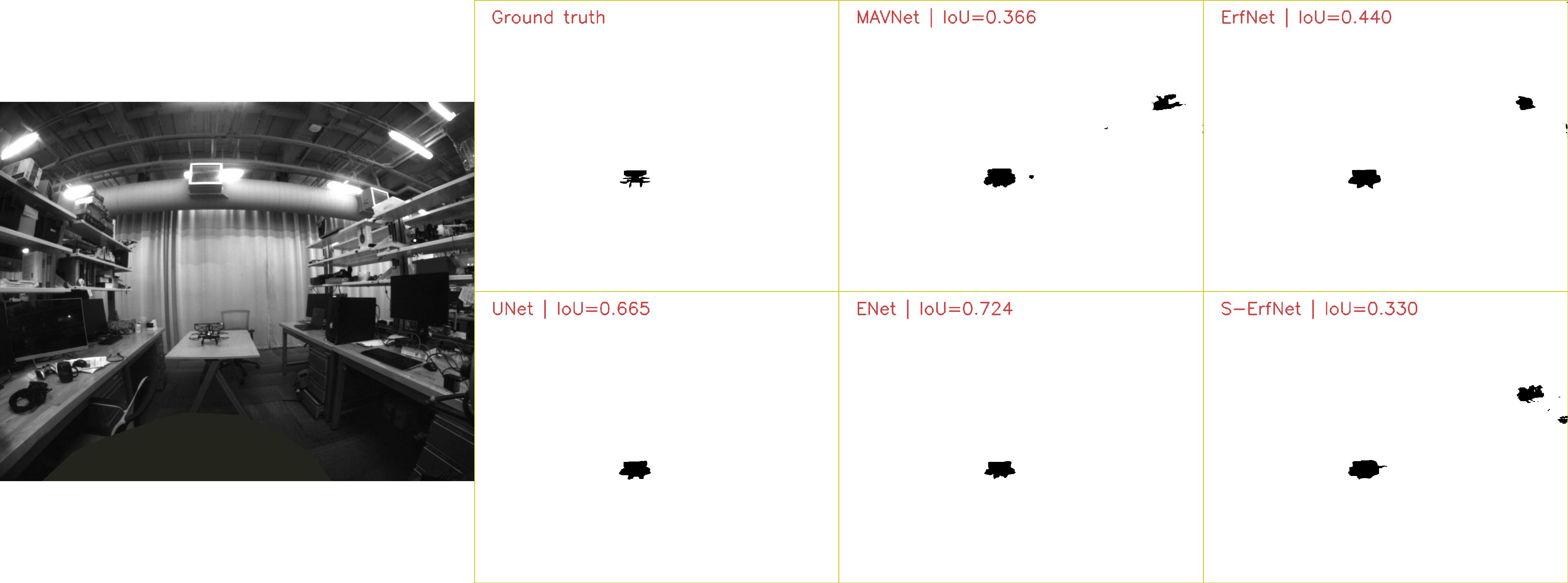}
    
    \caption{Failure case. MAVNet, ErfNet and S-ErfNet have high FP rate. Colors: black - MAV, white - background}
    \label{fig:drone_failure}
\end{figure*}

\begin{figure*}[t!]
    \vspace{3mm}
    \centering
    \includegraphics[width=0.8\textwidth]{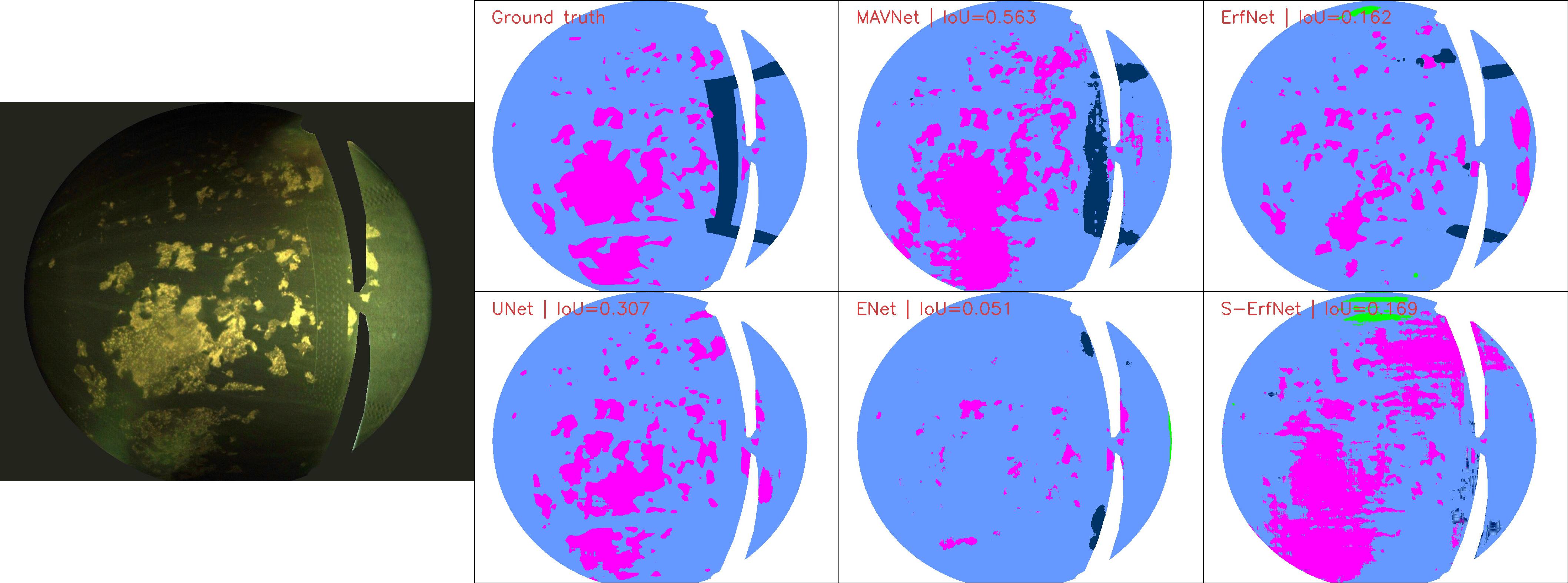} 
    
    \caption{Success case of MAVNet on the penstock dataset. Colors: pink - corrosion, blue - rivet, green - water.}
    \label{fig:penstock_qualitative_result}
\end{figure*}

Examples of success case and failure case of MAVNet on the MAVNet can be seen in Fig.~\ref{fig:drone_qualitative_result} and  Fig.~\ref{fig:drone_failure}, respectively. Fig.~\ref{fig:penstock_qualitative_result} illustrates a success case of MAV on the penstock dataset. In all examples, it can be seen that MAVNet yields a high number of FP pixels. UNet fails to detect the MAV in one case; ENet performs well on both MAV detection cases; ErfNet and S-ErfNet yield high numbers of FP pixels. These qualitative results are consistent with the quantitative results in Tab.~\ref{tab:drone_performance} and Tab.~\ref{tab:penstock_performance}. 

The performance difference of MAVNet regarding IoU on two datasets can be explained by the difference in $TP/TN$ between the two datasets. In the penstock dataset, total $TP/TN \sim 4:6$ while in the MAV dataset, $TP/TN \sim 1:200$.

\section{Conclusions}
\label{sec:conclusions}
In this work, we develop a fast and lightweight semantic segmentation model to satisfy SWaP constraints. We provide two datasets representing specific real-world tasks for autonomous MAVs. Compared to other models, MAVNet has a good tradeoff between inference time and performance on both datasets. Experiments show that a model can perform well on a dataset, with respect to one evaluation metric while performing poorly on the other datasets with other metrics. Our work demonstrates a potential for further work on designing modestly-sized networks with a manageable number of parameters to perform MAV tasks under SWaP constraints.  

\bibliographystyle{bib/IEEEtran}
\bibliography{bib/bib.bib}

\end{document}